%% file: emnlp2020.tex
\documentclass[11pt,a4paper]{article}
\usepackage[hyperref]{emnlp2020}
\usepackage{times}
\usepackage{latexsym, amsbsy}

\usepackage{graphicx}
\usepackage{booktabs} 

\usepackage{microtype}
\usepackage{enumitem} 
\usepackage{arydshln}

\include{definitions}

\aclfinalcopy 

\title{Think before you act: A simple baseline for compositional generalization}

\author{Christina Heinze-Deml\\
    Seminar for Statistics, ETH Zurich\\
    \texttt{heinzedeml@stat.math.ethz.ch}\\
  \\\And
  Diane Bouchacourt \\
  Facebook AI Research \\
  \texttt{dianeb@fb.com} \\}

\date{}

\makeatletter
\newcommand{\skipitems}[1]{%
  \addtocounter{\@enumctr}{#1}%
}
\makeatother

\begin{document}
\maketitle
\begin{abstract}
Contrarily to humans who have the ability to recombine familiar expressions to create novel ones, modern neural networks struggle to do so. This has been emphasized recently with the introduction of the benchmark dataset ``gSCAN'' \citep{Ruis2020}, aiming to evaluate models' performance at compositional generalization in grounded language understanding. 
In this work, we challenge the gSCAN benchmark by proposing a simple model that achieves surprisingly good performance on two of the gSCAN test splits. Our model is based on the observation that, to succeed on gSCAN tasks, the agent must (i) identify the target object (\emph{think}) before (ii) navigating to it successfully (\emph{act}). Concretely, we propose an attention-inspired modification of the baseline model from \citet{Ruis2020}, together with an auxiliary loss, that takes into account the sequential nature of steps (i) and (ii).  
While two compositional tasks are trivially solved with our approach, we also find that the other tasks remain unsolved, validating the relevance of gSCAN as a benchmark for evaluating models' compositional abilities. 
\end{abstract}

\section{Introduction}
While deep neural networks excel at tasks where training and test distributions are identical \citep{LeCun:etal:2015}, they often fail to generalize to test distributions that are in some metric close to, but different from the one seen during training \citep{Lake:etal:2016, HeinzeDeml2017, Arjovsky2019InvariantRM}. In the field of natural language understanding, there has been considerable focus on the lack of ``compositional generalization'', i.e.\ the phenomenon that models often cannot recombine familiar expressions from known parts \citep{Lake:Baroni:2017,Dessi:Baroni:2019,Loula:etal:2018}. The problem of designing models that exhibit human-like compositional skills is of utmost importance if they are to be deployed in real-world systems where robustness and trustworthiness are crucial. This has led to the design of benchmarks to assess the behavior and weaknesses of current state-of-the-art models. In this work, we focus on the benchmark gSCAN \citep{Ruis2020}. Our intuition is that, to succeed in a grounded environment, an agent must sequentially understand and act on its environment---as a human would. Thus, we ask the following question: by simply encouraging the model to ``think before acting'', can we solve some of the gSCAN challenges?

\section{Related Work}\label{sec:related_work}
\begin{figure}[t]
\vskip 0.in
\begin{center}
\centerline{\includegraphics[width=0.8\columnwidth]{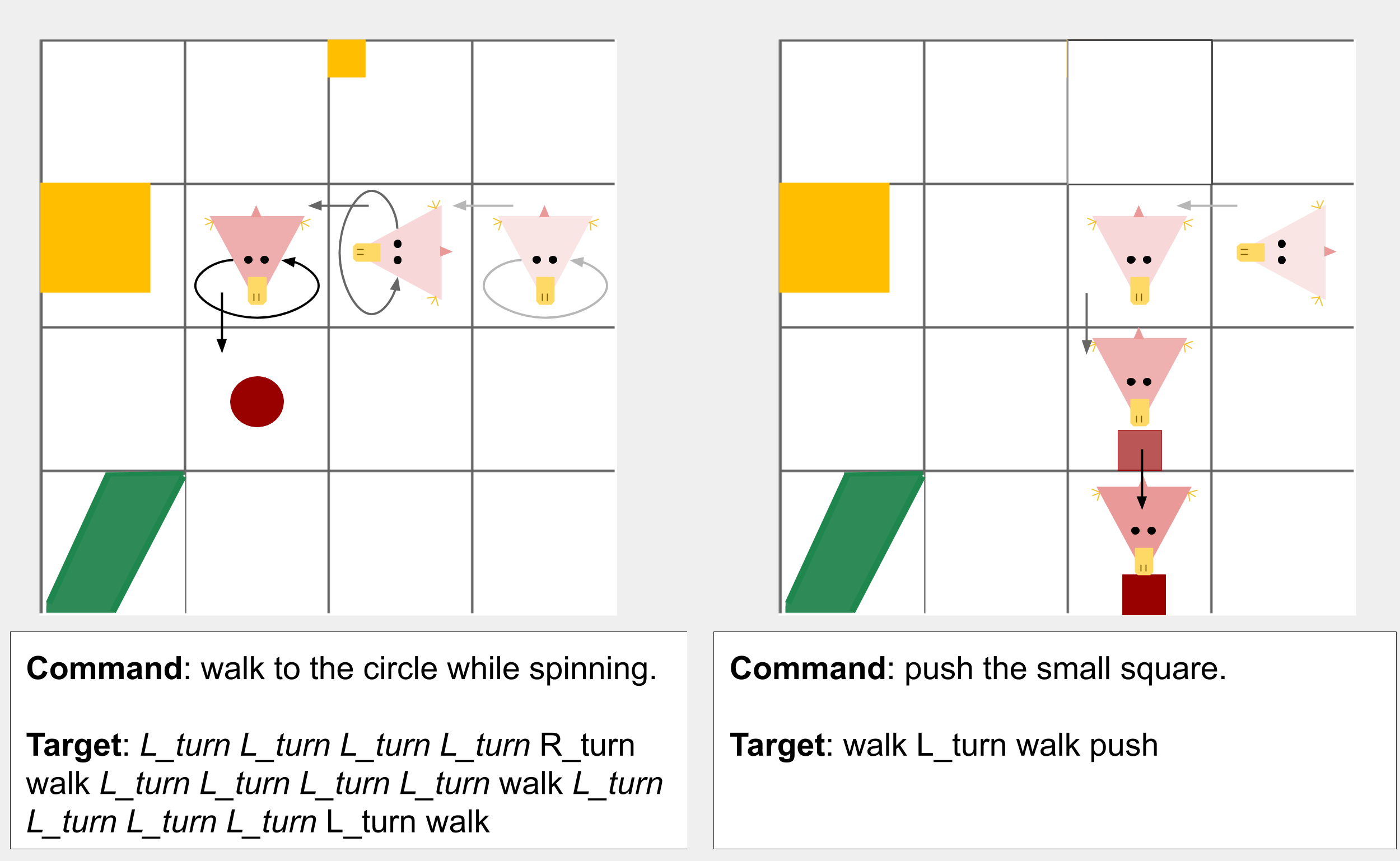}}
\caption{Exemplar commands, grid world and target action sequences in gSCAN. Image is taken from \citet{Ruis2020} with permission.}
\label{fig:gscan-example}
\end{center}
\vskip -0.3in
\end{figure}

Recently, various benchmark datasets for compositional generalization have been proposed \citep{Johnson:etal:2017, Lake:Baroni:2017, ChevalierBoisvert:etal:2019, Ruis2020}.
The gSCAN benchmark builds on SCAN \cite{Lake:Baroni:2017} which was designed to test for linguistic generalization and the ability to learn compositional rules. 
SCAN mimics a navigation environment by pairing linguistic commands (e.g.\ ``walk left'') with action sequences (L\_TURN, WALK). Training and test splits contain systematic differences. 
Recent work has made some progress on SCAN \citep{Gordon2020,Russin2019,Andreas:2019,LakeMeta2019} through architectural changes or data augmentation schemes. 
One limitation of SCAN is that it is not grounded in the visual world. This stands in contrast to real languages which are understood in relation to the state of the world. E.g., the meaning of adjectives may differ, depending on the objects present in the given scene. 

gSCAN \citep{Ruis2020} addresses this shortcoming by pairing input commands with the state of a grid world in which the agent is situated. For instance, the input may consist of the command ``Walk to the big square'' along with a representation of the grid world, containing a big square, several so-called distractor objects with different properties and the agent itself. The objects differ in shape, color and/or size. The task for the agent is to navigate from its current position in the grid world to the target object; see Figure~\ref{fig:gscan-example}. The construction of gSCAN allows for evaluating the ability of a trained agent to succeed at compositional challenges of different types. In total, there are 8 different test splits. \citet{Ruis2020} train and evaluate a baseline sequence-to-sequence (seq2seq) model and the GECA model \citep{Andreas:2019}, showing that both fail on most of the gSCAN tasks. 

In the context of Visual Question Answering, progress has been made by explicitly reasoning in a sequential manner \citep{Marois:etal:2018,Hudson:Manning:2018}. Our work follows a similar intuition as we explain below.

\section{Method}\label{sec:method}
In this section, we describe our modification of the multi-modal sequence-to-sequence model used in \citet{Ruis2020} that results in good performance on two gSCAN test splits. 
Our idea is motivated by the following reasoning. To succeed on all gSCAN test splits, the agent must ``think before acting'', i.e.\ it must be able to (i) identify the target object first, and (ii) navigate to it successfully. 
For some splits the required generalization is only in (i)---after having identified the target successfully, the navigation aspect is equivalent to the training environment. We describe these splits briefly below; in the original gSCAN paper these are referred to as (B), (C), and (E). For other splits, generalization in (ii) is needed as these splits affect \emph{how} the agent has to navigate to the target. 

\begin{enumerate}[itemsep=.5ex, leftmargin=.5cm, label=(\Alph*), topsep=.1ex, parsep=.5ex, partopsep=.5ex]
    \item \emph{Random split} This test split has the same distribution as the training set and serves to evaluate the agent's generalization skills when there is no distributional shift between training and testing. 
    \item \emph{Yellow squares} The training set contains examples with yellow squares as the target object but the color is never specified in the command, i.e.\ it may be referred to by ``the square'', ``the big square'', or ``the small square''. At test time, yellow squares are referred to as ``the yellow square'', ``the big yellow square'', or ``the small yellow square''. The model thus needs to generalize with respect to the target's references. 
    \item \emph{Red squares} This split assesses compositional skills with respect to attributes. In the training set red squares are never the target object. At test time, red squares are the target and the agent must identify it, based on having been trained with other squares and other red objects as target. 
    \skipitems{1}
    \item \emph{Relativity} To perform well on this split, the model needs to learn that size adjectives are relative to the other objects being present in the grid world. During training, circles of size 2 are never referred to as ``small''. They may be referred to without a size adjective or as ``big'', if there is another circle of size 1 present. At test time, circles of size 2 are paired with other larger circles and are referred to as ``small''. 
\end{enumerate}
Motivated by the fact that splits (B), (C) and (E) only require generalization in step (i), we hypothesize that adding the task of predicting the target object location improves overall performance. Indeed, \citet{Ruis2020} already considered adding the classification of the target object location as an auxiliary loss to the objective. As this attempt did not yield clear improvements, as a first step, we modify the auxiliary architecture that predicts the position of the target. Contrarily to \citet{Ruis2020}, we follow the principle of ``think before you act'' by first classifying the target from the encodings and decoding subsequently. However, as we show in an ablation study, this auxiliary loss alone is not sufficient to improve performance. Additionally, we incorporate the log-softmax scores of the target prediction task into the attention mechanism of the decoder as we detail below\footnote{This modification is not applicable to the auxiliary task architecture from \citet{Ruis2020}.}. Code of our modifications, based on the implementation by the gSCAN authors\footnote{\url{https://github.com/LauraRuis/multimodal_seq2seq_gSCAN}}, will be made available.

We now detail our proposed changes to the seq2seq baseline model from \citet{Ruis2020}. 
The input to the agent consists of the command sequence $\xt^c$ and the grid world representation $\Xt^s$. The original baseline model of \citet{Ruis2020} is composed of a command encoder $\mathbf{h}^c = f_c(\xt_c)$, a convolutional neural network (CNN) encoder of the grid world $\Ht^s = f_s(\Xt^s)$, and a decoder. 
The decoder operates on context vectors for the command ($\ct_j^c$) and the grid world ($\ct_j^s$), the embedding of the previous output symbol $\et_j^d$ and the previous decoder state $\mathbf{h}^d_{j-1}$. The context vector $\ct_j^c$ is computed using attention on the command encoding, $\ct_j^c = \mathrm{Attention}(\mathbf{h}^d_{j-1}, \mathbf{h}^c)$ , $\ct_j^s$ is computed with conditional attention over the world state features, $\ct_j^s = \mathrm{Attention}([\ct^c_j; \mathbf{h}^d_{j-1}], \mathbf{H}^s)$.  Supplementary material Figure \ref{fig:laura-model} describes the architecture used in \citet{Ruis2020}. Training is done by minimizing cross-entropy and supervised with ground-truth target sequences. We refer the reader to \citet{Ruis2020} for details on the training procedure. Optionally, the model proposed by \citet{Ruis2020} can be trained with the auxiliary task of predicting the location of the target object in the grid world with cross entropy loss and target position ground-truth. \citet{Ruis2020} simply use a log-softmax over the attention weights of each grid cell, summed over all decoder steps.

Our idea is to first help the model better focus on the target by modifying the way the target position is predicted in the auxiliary task. While predicting the target position is a classification task and thus does not require a recurrent architecture, we still get inspiration from sequential attention mechanisms and propose two new options:
\begin{enumerate}[itemsep=.1ex, leftmargin=.5cm, label=(\roman*), topsep=.1ex, parsep=.5ex, partopsep=.1ex]
    \item ``World" (relies mainly on world state features): we predict the target position based on the last hidden state of the command encoder $h^c_n$ and attention-weighted world encodings. We use scaled dot-product attention \citep{vaswani2017} between $h^c_n$ and the world encodings, $\tilde{\mathbf{H}}^s = \mathrm{DotAttention}(h^c_n, \Ht^s)$ and concatenate it with $h^c_n$ in a vector $\vt =[\tilde{\mathbf{H}}^s; h^c_n]$.
    \item ``Both": we predict the target position based on attention-weighted command encodings and attention-weighted world encodings. First, we compute $\tilde{\mathbf{h}}^c = \mathrm{DotAttention}(\Ht^s, \mathbf{h}^c)$ to yield weighted command encodings and compute their weighted sum. Second, we compute $\tilde{\Ht}^s = \mathrm{DotAttention}(\tilde{\mathbf{h}}^c, \Ht^s)$ to yield weighted world encodings and concatenate them as $\vt = [\tilde{\mathbf{H}}^s; \tilde{\mathbf{h}}^c]$.
\end{enumerate}
In both options, the concatenated vector $\vt$ is fed to a linear classification layer that predicts scores for each position to be the target position. 
Introducing the attention mechanism for the target prediction task as described above only constrains the encoders. Thus, we additionally weigh the world encodings $\Ht^s$ by the log-softmax scores of the target position prediction linear layer before feeding them to the attention mechanism that computes the context vector $\ct^s_j$. This also constrains the decoder.


\begin{table*}[!t]
		\centering
		\setlength\tabcolsep{2.5pt} 
		\caption{Percentage of observations for which the target output sequence was predicted exactly. Results for Baseline w/o aux (refers to the baseline model from \citet{Ruis2020}) and GECA are taken from \citet{Ruis2020}.}
		\begin{tabular}{lc|c|c|c|c}
            \textbf{}                                    & \multicolumn{5}{c}{\textbf{Exact Match (\%)}}                                   \\ 
            \multicolumn{1}{l}{\textbf{Split}}         & \multicolumn{1}{c}{\textbf{Baseline w/o aux}} & \multicolumn{1}{c}{\textbf{GECA}} & \multicolumn{1}{c}{\textbf{Baseline w/ aux}} & \multicolumn{1}{|c}{\textbf{Ours (world)}} & \multicolumn{1}{c}{\textbf{Ours (both)}}\\
            \multicolumn{1}{l}{A: Random} & $\mathbf{97.69 \pm 0.22}$                          & $87.6 \pm 1.19$         
                & $92.88 \pm 0.78$ & $93.94 \pm 0.7$ & $94.19 \pm 0.71$\\ \hline
            \multicolumn{1}{l}{B: Yellow squares} & $54.96 \pm 39.39$                 & $34.92 \pm 39.30$ 
                & $61.18 \pm 30.1$ & $\mathbf{87.31 \pm 4.38}$ & $86.45 \pm 6.28$\\ \hline
            \multicolumn{1}{l}{C: Red squares}& $23.51 \pm 21.82$                     & $78.77 \pm 6.63$   
                & $11.73 \pm 5.05$ & $79.77 \pm 11.01$ & $\mathbf{81.07 \pm 10.12}$\\ \hline
            \multicolumn{1}{l}{E: Relativity}& $35.02 \pm 2.35$                       & $33.19 \pm 3.69$    
                & $50.51 \pm 14.08$ & $\mathbf{52.8 \pm 9.96}$ & $43.43 \pm 7.0$\\ 
            \end{tabular}
            \label{table:results}
\end{table*}


\section{Experiments}\label{sec:experiments}
We compare the performance of our proposals ``world'' and ``both'' with the results reported in \citet{Ruis2020} as well as their suggestion for training with the auxiliary loss. Following \citet{Ruis2020}, we train the model for $200 000$ iterations. Most hyperparameters are identical; we performed hyperparameters tuning on the dropout rates of the encoders, the decoder and the weight of the auxiliary task in the training objective. Each setting is run five times over different random seeds. We report mean and standard errors of the percentage of exact matches, i.e.\ the percentage of observations for which the target output sequence is predicted \emph{exactly}. Further details can be found in Section~\ref{subsec:add_results}.

Table~\ref{table:results} summarizes the results. The first two columns (``Baseline w/o aux'' and ``GECA'') are taken from \citet{Ruis2020}. ``Baseline w/ aux'' shows the results obtained when training with the auxiliary loss but using the proposal from \citet{Ruis2020} described above. The results in the last two columns are obtained with our proposed modifications. First, we observe that all models obtain close-to-perfect performance on the \emph{Random} test split. Hence all models learn the navigation task successfully when there are no systematic differences between training and test split. Second, we see that ``Baseline w/ aux'' does not improve upon ``Baseline w/o aux'' reliably for splits (B), (C) and (E)---even though mean performance is higher for (B) and (E), the standard errors are very large. Our proposals outperform all other models on \emph{Yellow squares} with larger mean performances and smaller standard errors. For \emph{Red squares}, ``world'' and ``both'' outperform the two baselines and achieve comparable performance as GECA. Importantly, GECA regenerates the grid world to match the augmented commands. Hence, it uses more supervision than our proposals. For \emph{Relativity}, ``world'' and ``both'' outperform ``Baseline w/o aux'' and GECA while performing similarly as ``Baseline w/ aux''. While our method improves performance largely on splits (B) and (C), performance on split (E) is still unsatisfactory with exact match of at most $\approx 50\%$.  

\textbf{Error analysis}
First, we would like to analyze to what extent target prediction accuracy correlates with exact match performance as reported in Table~\ref{table:results}. This analysis sheds light on whether there are cases where the target is identified successfully and hence compositional generalization is achieved while exact match performance is poor due to decoding issues only. Supplementary Table~\ref{table:results_target_pred} in Section~\ref{subsec:add_results} contains the target prediction accuracies for all models with auxiliary loss; Figure~\ref{fig:results} plots the mean exact match performances vs.\ mean target prediction accuracies. For our models, we observe a large correlation between the performances and there are no indications for the described issue. While exact match accuracy is always somewhat lower, this is to be expected due to the larger degree of difficulty of the task.

Second, Table~\ref{table:results} shows that while training with the auxiliary task leads to improvements on split (E), these are much smaller than for splits (B) and (C). For split (E), ``world'' and ``both'' do not improve upon ``Baseline w/ aux'' in terms of exact match; improvements in terms of target prediction accuracy are only marginal (for ``world''). When looking at the performance statistics by referred target, we find that the model performs much better when the color is specified (see Table~\ref{table:results_E_ref_target}). In these cases, the agent can exploit the fact that there are only two circles of the specific color in the grid world\footnote{This observation was also made in \citet{Ruis2020}.}. When color is not specified, the agent needs to choose from a typically larger number of circles. Indeed, performance is much lower in these cases ($\approx 28-42\%$). This suggests that the agent genuinely fails to understand relativity which seems considerably more difficult than the required generalization in splits (B) and (C). While (B) and (C) require the agent to understand novel combinations of known parts, in (E) it needs to connect the \emph{same} command specifying a particular attribute (here, size: ``the small circle'') with objects having \emph{different} values of that property (``small'' circles may be of size 1, 2 or 3). 

Finally, we evaluate the performance of an ablated version of our model where we do not weigh the world encodings $\Ht_s$ by the log-softmax scores of the target position prediction task. Supplementary Section~\ref{subsec:add_results} shows that while the ablated version performs well in terms of target prediction accuracy, exact match performance is much lower than the full model performance from Table~\ref{table:results}.


\section{Discussion}
The gSCAN test splits are well-designed as each test split tackles generalization either in (i) target identification, or (ii) navigation. Our hypothesis was that performance on tasks which only require generalization at (i) can be improved by improving performance on target location prediction, and proposed a sequential method that relied on the ``think before you act" principle. Our hypothesis was confirmed for splits (B) and (C) that require understanding \emph{new combinations of known features}. 
Interestingly, we also found that the \emph{Relativity} split (E) remains unsolved. Even though adding the auxiliary loss improves performance, our ``think before you act" modification does not yield further improvements in contrast to (B) and (C), neither in terms of target prediction nor exact match accuracy. It hence remains an open question how to tackle target prediction in the context of relativity, and whether this would lead to corresponding performance gains for the full task. 

Overall, our results emphasize the relevance of gSCAN as a benchmark for evaluating models' compositional abilities, especially as generalization in terms of navigation remains an open challenge. 
\clearpage
\newpage
\bibliographystyle{acl_natbib}
\bibliography{anthology,emnlp2020,marco,additional,library_clean}

\appendix
\clearpage
\newpage

\renewcommand{\thetable}{A.\arabic{table}}
\renewcommand{\thefigure}{A.\arabic{figure}}
\section{Appendices}
\label{sec:appendix}

\begin{figure*}[h]
	\centering
	\includegraphics[width=1.\textwidth]{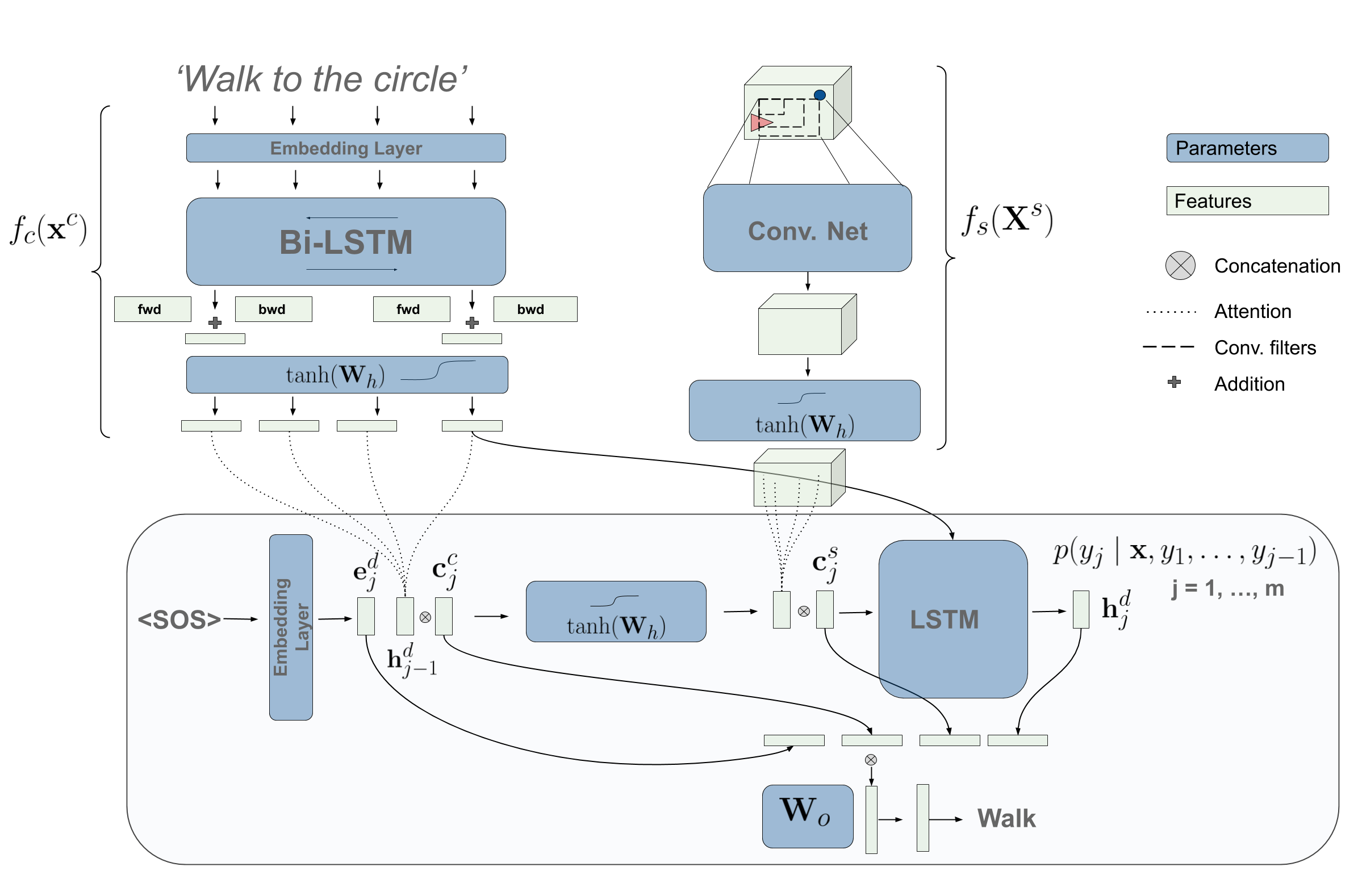}
	\captionof{figure}{Baseline neural network from \citet{Ruis2020}. The command encoder parses the input command using a biLSTM $f_c$, and the state encoder parses the grid world state with the CNN $f_s$. An LSTM decoder outputs the predicted action sequence (bottom), using joint attention over the command and the world state. Image is taken from \citet{Ruis2020} with permission.}
	\label{fig:laura-model}
\end{figure*}

\subsection{Additional experimental details and results}\label{subsec:add_results}

\paragraph{Experimental details} The gSCAN dataset is available from \url{https://github.com/LauraRuis/groundedSCAN}.
We select the best hyperparameter setting by choosing the best exact match result on the provided \emph{Dev} split, which is a small subset of the \emph{Random} test split, both having the same distribution as the training set. We experiment with command encoder and sequence decoder dropout rates $\in [0,0.3]$, world state encoder dropout rate $\in [0,0.3]$, and weight of the auxiliary task $\in [0.3,0.7]$. We leave all other hyperparameters as in \citet{Ruis2020}. Code of our modifications, based on the implementation by the gSCAN authors\footnote{\url{https://github.com/LauraRuis/multimodal_seq2seq_gSCAN}}, will be made available.

In the two proposed modifications of the target location prediction task ``world" and ``both", the concatenated vector $\vt$ is of size $d^2*3*c_{out}+h_e$ as the original architecture CNN encoder uses $3$ different kernel sizes. $c_{out}$ is the number of features maps per kernel size, $h_e$ is the command encoder hidden size, and $d\times d$ is the total number of cells in a grid world of width and height $d$. Accordingly, the linear layer that performs target position prediction is of input size $d^2*3*c_{out}+h_e$ and output size $d^2$.

We report results on gSCAN test splits (B), (C) and (E) as no improvements were obtained on the remaining splits. As those splits require generalization in \emph{how} the agent has to navigate to the target, we also do not expect obtaining improvements on these splits with our proposed methodology. Hence, it remains on open question how to make progress on these. 

\paragraph{Best hyperparameters chosen with cross-validation for performance shown in Tables \ref{table:results}, \ref{table:results_target_pred} and \ref{table:results_E_ref_target}}~\\
\emph{Baseline w/ aux}:
cnn dropout rate: 0.1, decoder dropout rate: 0.0, encoder dropout rate: 0.0, weight auxiliary task: 0.3 \\
\emph{World}:
cnn dropout rate: 0.0, decoder dropout rate: 0.0, encoder dropout rate: 0.3, weight auxiliary task: 0.3
\\
\emph{Both}:
cnn dropout rate: 0.0, decoder dropout rate: 0.0, encoder dropout rate: 0.3, weight auxiliary task: 0.7

\paragraph{Best hyperparameters chosen with cross-validation for ablated performance shown in Tables \ref{table:results_abalation}}~\\
\emph{World}:
cnn dropout rate: 0.0, decoder dropout rate: 0.0, encoder dropout rate: 0.3, weight auxiliary task: 0.3
\\
\emph{Both}:
cnn dropout rate: 0.1, decoder dropout rate: 0.3, encoder dropout rate: 0.3, weight auxiliary task: 0.3

\begin{figure*}[h]
	\centering
	\includegraphics[width=1.\textwidth]{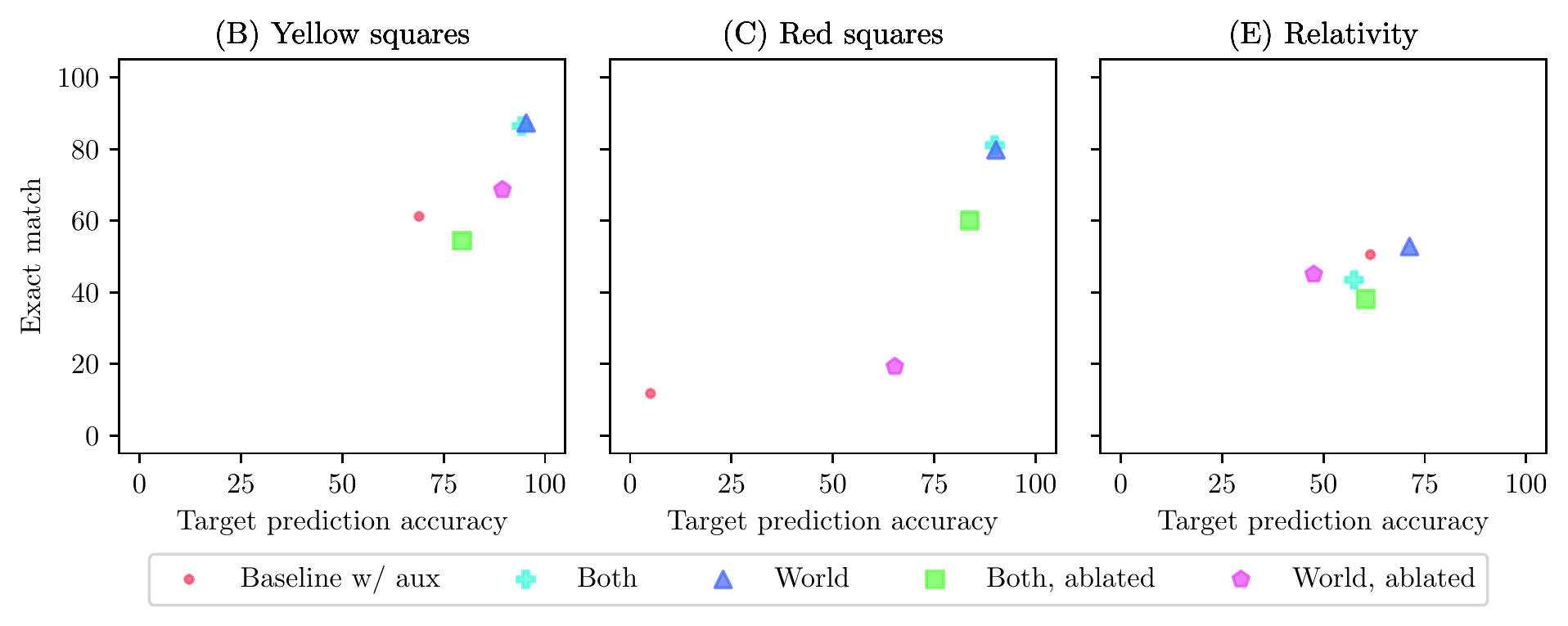}
	\captionof{figure}{Exact match averages vs.\ mean target prediction accuracy, for splits (B), (C) and (E). Splits (B) and (C): For ``world'' and ``both'', we observe a large correlation between exact match and target prediction performance. For the ablated models ``world, ablated'' and ``both, ablated'' exact match accuracies are much lower, despite partially good target prediction accuracies. Split (E) remains challenging for all models with exact match accuracies of at most $\approx 50\%$.}
	\label{fig:results}
\end{figure*}

\paragraph{Additional experimental results}
Table~\ref{table:results_target_pred} shows the target prediction accuracies for the models with auxiliary loss and  Figure~\ref{fig:results} plots the mean exact match performances vs.\ mean target prediction accuracies. Overall, we observe a large correlation between the two performance statistics for our proposed models. Table~\ref{table:results_E_ref_target} breaks down the exact match accuracies for split (E) by the referred target. Performance is much lower when no color attribute is specified. 

\paragraph{Ablation study}
One may ask whether the classification from the attention-weighted command and state encodings $\vt$ alone is sufficient to achieve the improvements reported in Section~\ref{sec:experiments}. Table~\ref{table:results_abalation} contains the results obtained in this fashion, i.e.\ without weighing the world encodings $\Ht_s$ by the log-softmax scores of the target position prediction task. Comparing Tables~\ref{table:results} and~\ref{table:results_abalation}, as well as Figure~\ref{fig:results}, shows that weighing the world encodings $\Ht_s$ by the log-softmax scores of the target position prediction task is indeed beneficial: For all splits, the exact match accuracies of the ablated models are lower. While exact match performance is worse for the ablated models, we observe that target prediction accuracy is comparable on split (B) for ``world'' and on split (C) for ``both'' (see Tables~\ref{table:results_target_pred} and~\ref{table:results_abalation}; Figure~\ref{fig:results}). This suggests that the weighing the world encodings $\Ht_s$ by the log-softmax scores of the target position prediction task indeed yields the desired changes in the decoder, needed to obtain the improved exact match performances reported in Section~\ref{sec:experiments}.

\begin{table*}[!t]
		\centering
		\caption{Target prediction accuracies for models with auxiliary loss}
		\begin{tabular}{l|c|c|c|}
            \cline{2-4}
            \textbf{}                                    & \multicolumn{3}{c|}{\textbf{Target prediction accuracy (\%)}}                                   \\ \hline
            \multicolumn{1}{|l|}{\textbf{Split}}         &  \multicolumn{1}{c|}{\textbf{Baseline w/ aux}} & \multicolumn{1}{c|}{\textbf{Ours (world)}} & \multicolumn{1}{c|}{\textbf{Ours (both)}}\\ \hline \hline
            \multicolumn{1}{|l|}{A: Random} & $95.5 \pm 2.51$ & $100.0 \pm 0.00$ & $100.0 \pm 0.00$   \\ \hline \hline
            \multicolumn{1}{|l|}{B: Yellow squares} & $68.94 \pm 36.6$ & $95.36 \pm 4.26$ & $94.24 \pm 5.7$ \\ \hline
            \multicolumn{1}{|l|}{C: Red squares}& $5.02 \pm 6.8$ & $90.23 \pm 10.35$ & $89.9 \pm 9.7$ \\ \hline
            \multicolumn{1}{|l|}{E: Relativity}& $61.58 \pm 19.36$ & $71.26 \pm 13.41$ & $57.57 \pm 12.15$ \\ \hline
            \end{tabular}
            \label{table:results_target_pred}
\end{table*}


\begin{table*}[!t]
		\centering
		\caption{Split (E): Exact match performance by referred target}
		\begin{tabular}{l|c|c|c|}
            \cline{2-4}
            \textbf{}                                    & \multicolumn{3}{c|}{\textbf{Exact match (\%)}}                                   \\ \hline
            \multicolumn{1}{|l|}{\textbf{Referred target}}         &  \multicolumn{1}{c|}{\textbf{Baseline w/ aux}} & \multicolumn{1}{c|}{\textbf{Ours (world)}} & \multicolumn{1}{c|}{\textbf{Ours (both)}}\\ \hline \hline
            \multicolumn{1}{|l|}{small yellow circle}& $66.5 \pm 12.44$& $69.62 \pm 11.72$& $57.2 \pm 11.44$\\ \hline
            \multicolumn{1}{|l|}{small green circle}& $65.57 \pm 11.92$& $64.46 \pm 6.21$& $62.62 \pm 15.47$\\ \hline
            \multicolumn{1}{|l|}{small red circle}& $66.72 \pm 11.65$& $57.59 \pm 11.38$& $54.08 \pm 3.54$\\ \hline
            \multicolumn{1}{|l|}{small blue circle}& $65.99 \pm 10.36$& $61.72 \pm 9.11$& $59.25 \pm 5.69$\\ \hline
            \multicolumn{1}{|l|}{small  circle}& $34.83 \pm 16.75$& $42.24 \pm 15.24$& $28.57 \pm 9.45$\\ \hline

            \end{tabular}
            \label{table:results_E_ref_target}
\end{table*}


\begin{table*}[!t]
		\centering
		\caption{Exact match and target prediction accuracy for ablated models: Here, target location prediction is based on the attention-weighted command and state encodings $\vt$ but the weighing of the world encodings $\Ht_s$ by the log-softmax scores of the target position prediction task is omitted.}
		\begin{tabular}{l|c|c||c|c|}
            \cline{2-5}
            \textbf{}                                    & 
            \multicolumn{2}{c||}{\textbf{Exact match (\%)}} &
            \multicolumn{2}{c|}{\textbf{Target prediction accuracy (\%)}} \\ \hline
            \multicolumn{1}{|l|}{\textbf{Split}}         &   
            \multicolumn{1}{c|}{\textbf{World, ablated}} & 
            \multicolumn{1}{c||}{\textbf{Both, ablated}} &
            \multicolumn{1}{c|}{\textbf{World, ablated}} & 
            \multicolumn{1}{c|}{\textbf{Both, ablated}}\\ \hline \hline
            \multicolumn{1}{|l|}{A: Random} & $89.98 \pm 3.04$ & $93.15 \pm 0.35$ & $100.0 \pm 0.00$ & $100.0 \pm 0.00$ \\ \hline \hline
            \multicolumn{1}{|l|}{B: Yellow squares} & $68.65 \pm 11.51$ & $54.38 \pm 17.14$ & $89.5 \pm 3.58$& $79.53 \pm 24.04$ \\ \hline
            \multicolumn{1}{|l|}{C: Red squares}& $19.23 \pm 13.53$ & $60.03 \pm 12.91$ & $65.3 \pm 22.66$ & $83.72 \pm 21.19$ \\ \hline
            \multicolumn{1}{|l|}{E: Relativity}& $45.03 \pm 10.77$ & $38.11 \pm 4.11$ & $47.62 \pm 8.39$ & $60.45 \pm 13.04$ \\ \hline
            \end{tabular}
            \label{table:results_abalation}
\end{table*}

\end{document}

%% file: definitions.tex
\newcommand{\ct}{\mathbf{c}}
\newcommand{\et}{\mathbf{e}}
\newcommand{\xt}{\mathbf{x}}
\newcommand{\vt}{\mathbf{v}}

\newcommand{\Ht}{\mathbf{H}}
\newcommand{\Xt}{\mathbf{X}}

%% file: emnlp2020.bbl
\begin{thebibliography}{17}
\expandafter\ifx\csname natexlab\endcsname\relax\def\natexlab#1{#1}\fi

\bibitem[{Andreas(2019)}]{Andreas:2019}
Jacob Andreas. 2019.
\newblock Measuring compositionality in representation learning.
\newblock In \emph{Proceedings of ICLR}, New Orleans, LA.
\newblock Published online:
  \url{https://openreview.net/group?id=ICLR.cc/2019/conference}.

\bibitem[{Arjovsky et~al.(2019)Arjovsky, Bottou, Gulrajani, and
  Lopez-Paz}]{Arjovsky2019InvariantRM}
Mart{\'i}n Arjovsky, L{\'e}on Bottou, Ishaan Gulrajani, and David Lopez-Paz.
  2019.
\newblock Invariant risk minimization.
\newblock \emph{ArXiv}, abs/1907.02893.

\bibitem[{{Chevalier-Boisvert} et~al.(2019){Chevalier-Boisvert}, Bahdanau,
  Lahlou, Willems, Saharia, Nguyen, and Bengio}]{ChevalierBoisvert:etal:2019}
Maxime {Chevalier-Boisvert}, Dzmitry Bahdanau, Salem Lahlou, Lucas Willems,
  Chitwan Saharia, Thien~Huu Nguyen, and Yoshua Bengio. 2019.
\newblock {BabyAI}: A platform to study the sample efficiency of grounded
  language learning.
\newblock In \emph{Proceedings of ICLR}, New Orleans, LA.
\newblock Published online:
  \url{https://openreview.net/group?id=ICLR.cc/2019/conference}.

\bibitem[{Dess\`{i} and Baroni(2019)}]{Dessi:Baroni:2019}
Roberto Dess\`{i} and Marco Baroni. 2019.
\newblock {CNNs} found to jump around more skillfully than {RNNs}:
  Compositional generalization in seq2seq convolutional networks.
\newblock In \emph{Proceedings of ACL}, Firenze, Italy.
\newblock {I}n press.

\bibitem[{Gordon et~al.(2020)Gordon, Lopez-Paz, Baroni, and
  Bouchacourt}]{Gordon2020}
Jonathan Gordon, David Lopez-Paz, Marco Baroni, and Diane Bouchacourt. 2020.
\newblock \href {https://openreview.net/forum?id=SylVNerFvr} {Permutation
  equivariant models for compositional generalization in language}.

\bibitem[{Heinze-Deml and Meinshausen(2017)}]{HeinzeDeml2017}
Christina Heinze-Deml and Nicolai Meinshausen. 2017.
\newblock Conditional variance penalties and domain shift robustness.
\newblock \emph{arXiv preprint arXiv:1710.11469}.

\bibitem[{Hudson and Manning(2018)}]{Hudson:Manning:2018}
Drew~A Hudson and Christopher~D Manning. 2018.
\newblock Compositional attention networks for machine reasoning.

\bibitem[{Johnson et~al.(2017)Johnson, Hariharan, van~der Maaten, Fei{-}Fei,
  Zitnick, and Girshick}]{Johnson:etal:2017}
Justin Johnson, Bharath Hariharan, Laurens van~der Maaten, Li~Fei{-}Fei,
  Lawrence Zitnick, and Ross Girshick. 2017.
\newblock {CLEVR:} a diagnostic dataset for compositional language and
  elementary visual reasoning.
\newblock In \emph{Proceedings of CVPR}, pages 1988--1997, Honolulu, HI.

\bibitem[{Lake and Baroni(2018)}]{Lake:Baroni:2017}
Brenden Lake and Marco Baroni. 2018.
\newblock Generalization without systematicity: On the compositional skills of
  sequence-to-sequence recurrent networks.
\newblock In \emph{Proceedings of ICML}, pages 2879--2888, Stockholm, Sweden.

\bibitem[{Lake et~al.(2017)Lake, Ullman, Tenenbaum, and
  Gershman}]{Lake:etal:2016}
Brenden Lake, Tomer Ullman, Joshua Tenenbaum, and Samuel Gershman. 2017.
\newblock Building machines that learn and think like people.
\newblock \emph{Behavorial and Brain Sciences}, 40:1--72.

\bibitem[{Lake(2019)}]{LakeMeta2019}
Brenden~M Lake. 2019.
\newblock \href {http://arxiv.org/abs/1906.05381} {{Compositional
  generalization through meta sequence-to-sequence learning}}.
\newblock In \emph{{Advances in Neural Information Processing Systems}}.

\bibitem[{LeCun et~al.(2015)LeCun, Bengio, and Hinton}]{LeCun:etal:2015}
Yann LeCun, Yoshua Bengio, and Geoffrey Hinton. 2015.
\newblock Deep learning.
\newblock \emph{Nature}, 521:436--444.

\bibitem[{Loula et~al.(2018)Loula, Baroni, and Lake}]{Loula:etal:2018}
Joao Loula, Marco Baroni, and Brenden Lake. 2018.
\newblock Rearranging the familiar: Testing compositional generalization in
  recurrent networks.
\newblock In \emph{Proceedings of the EMNLP BlackboxNLP Workshop}, pages
  108--114, Brussels, Belgium.

\bibitem[{Marois et~al.(2018)Marois, Jayram, Albouy, Kornuta, Bouhadjar, and
  Ozcan}]{Marois:etal:2018}
Vincent Marois, TS~Jayram, Vincent Albouy, Tomasz Kornuta, Younes Bouhadjar,
  and Ahmet~S Ozcan. 2018.
\newblock On transfer learning using a mac model variant.
\newblock \emph{arXiv preprint arXiv:1811.06529}.

\bibitem[{Ruis et~al.(2020)Ruis, Andreas, Baroni, Bouchacourt, and
  Lake}]{Ruis2020}
Laura Ruis, Jacob Andreas, Marco Baroni, Diane Bouchacourt, and Brenden~M.
  Lake. 2020.
\newblock A benchmark for systematic generalization in grounded language
  understanding.
\newblock \emph{ArXiv}, abs/2003.05161.

\bibitem[{Russin et~al.(2019)Russin, Jo, O'Reilly, and Bengio}]{Russin2019}
Jake Russin, Jason Jo, Randall~C. O'Reilly, and Yoshua Bengio. 2019.
\newblock \href {http://arxiv.org/abs/1904.09708} {{Compositional
  generalization in a deep seq2seq model by separating syntax and semantics}}.
\newblock \emph{arXiv preprint}.

\bibitem[{Vaswani et~al.(2017)Vaswani, Shazeer, Parmar, Uszkoreit, Jones,
  Gomez, Kaiser, and Polosukhin}]{vaswani2017}
Ashish Vaswani, Noam Shazeer, Niki Parmar, Jakob Uszkoreit, Llion Jones,
  Aidan~N Gomez, \L~ukasz Kaiser, and Illia Polosukhin. 2017.
\newblock Attention is all you need.
\newblock In \emph{Advances in Neural Information Processing Systems 30}.

\end{thebibliography}
